# CSFNet: A Cosine Similarity Fusion Network for Real-Time RGB-X Semantic Segmentation of Driving Scenes

Danial Qashqai[a,][*], Emad Mousavian[a], Shahriar B. Shokouhi[a], Sattar Mirzakuchaki[a]

[a]*Department of Electrical Engineering, Iran University of Science and Technology, Tehran, Iran*

**Abstract**

Semantic segmentation, as a crucial component of complex visual interpretation, plays a fundamental role in autonomous vehicle vision systems. Recent studies have significantly improved the accuracy of semantic segmentation by exploiting complementary information and developing multimodal methods. Despite the gains in accuracy, multimodal semantic segmentation methods suffer from high computational complexity and low inference speed. Therefore, it is a challenging task to implement multimodal methods in driving applications. To address this problem, we propose the Cosine Similarity Fusion Network (CSFNet) as a real-time RGB-X semantic segmentation model. Specifically, we design a Cosine Similarity Attention Fusion Module (CS-AFM) that effectively rectifies and fuses features of two modalities. The CS-AFM module leverages cross-modal similarity to achieve high generalization ability. By enhancing the fusion of cross-modal features at lower levels, CS-AFM paves the way for the use of a single-branch network at higher levels. Therefore, we use dual and single-branch architectures in an encoder, along with an efficient context module and a lightweight decoder for fast and accurate predictions. To verify the effectiveness of CSFNet, we use the Cityscapes, MFNet, and ZJU datasets for the RGB-D/T/P semantic segmentation. According to the results, CSFNet has competitive accuracy with state-of-the-art methods while being state-of-the-art in terms of speed among multimodal semantic segmentation models. It also achieves high efficiency due to its low parameter count and computational complexity. The source code for CSFNet will be available at https://github.com/Danial-Qashqai/CSFNet.

## 1. Introduction

    Semantic segmentation is a fundamental task in computer vision, dealing with the analysis and understanding of driving scenes through pixel-level classification. Due to the high sensitivity of Advanced Driver-Assistance Systems (ADAS) and the potential for serious accidents in the event of errors, improving the accuracy of semantic segmentation models is very important.

    Recent advancements in sensor technology and the availability of complementary data such as depth [1], thermal [2], and polarization [3] have opened new doors to the development of multimodal semantic segmentation models in driving scenarios. Research in the field of multimodal semantic segmentation [4], [5], by fusing complementary information and RGB images, enables a deeper understanding of content and provides higher accuracy than RGB models [6], [7]. This superiority in accuracy is accompanied by overcoming challenges such as similar color or texture of objects, lighting variations, limited visibility, and light reflection from glossy surfaces.

    In the development of multimodal semantic segmentation, four approaches—early fusion [8], mid-term fusion [9], late fusion [10], and multi-level interactive fusion [11], [12], [13]—have been used to combine two

\* Corresponding author.
 *E-mail address:* ghashghaie_danial@elec.iust.ac.ir.

input modalities. Early fusion involves fusing the input data before extracting features; such a simple fusion ignores the complementarity between the input information [14]. Therefore, other aforementioned approaches have been proposed to perform cross-modal fusion using a two-branch network. Among these methods, multi-level interactive fusion stands out as the leading approach for multimodal semantic segmentation models, achieving superior accuracy. In this approach, the extracted feature maps of the branches are fused at multiple levels. While two-branch networks and multi-level fusion can improve accuracy, they increase computational complexity and dramatically slow down inference speed. Therefore, given the importance of processing speed in driving applications, the use of multimodal models in this domain is challenging.

The process of performing fusion operations is a key aspect of multimodal semantic segmentation. Early works [11], [15] fuse cross-modal features in a straightforward manner by applying element-wise addition, overlooking the complementary nature of the features. Recent works [16], [17] have improved this problem by using attention-based fusion modules. These modules generally use global information either directly [18] or by applying interactions between them [19] in a trainable approach. Despite the improvement in accuracy, the global information does not adequately distinguish between the features of the two modalities.

To address these shortcomings, we propose a Cosine Similarity Fusion Network (CSFNet) for real-time RGB-X semantic segmentation. In this model, an optimized encoder extracts features from both RGB and X modalities. The proposed encoder employs a two-branch architecture for the first three levels and a single branch for the final two levels. This approach reduces computational complexity and results in higher processing speeds. In addition, we further achieve higher computational efficiency by using the Short-Term Dense Concatenate (STDC) [20] backbone for the first time in a multimodal semantic segmentation model. To more effectively combine the features of two modalities, we design the Cosine Similarity Attention Fusion Module (CS-AFM). As a novel approach, this module rectifies and fuses the modalities by employing the cosine similarity between the corresponding channels in an attention-based approach. Unlike previous methods, the CS-AFM module considers local features by applying average pooling layers, and it can effectively distinguish cross-modal features by exploiting cosine similarity. This module, with its high generalization ability, is used in the CSFNet model for both fusion of RGB-X features in the encoder and fusion of skip connection features with the decoder. Finally, the proposed encoder is combined with an efficient context module and a lightweight decoder for fast and accurate predictions.

We evaluate the proposed CSFNet model on three types of multimodal semantic segmentation tasks, including RGB-Depth, RGB-Thermal, and RGB-Polarization. Given the research focus on driving scenarios, we use the Cityscapes [1], MFNet [2], and ZJU [3] datasets in this evaluation. In general, CSFNet achieves competitive accuracy with state-of-the-art (SOTA) multimodal semantic segmentation models; it also has the fastest inference speed of all multimodal semantic segmentation models, and because of its low complexity, it can be used in embedded hardware.

The main contributions of this study are summarized as follows:

- We leverage both dual and single-branch architectures to design an optimized encoder network. Furthermore, the proposed encoder uses the STDC backbone for the first time in a multimodal semantic segmentation task.
- We propose a Cosine Similarity Attention Fusion Module (CS-AFM) that rectifies and fuses the input features based on cross-modal similarity.
- We propose a Cosine Similarity Fusion Network (CSFNet) as a real-time RGB-X semantic segmentation model.
- CSFNet achieves competitive accuracy on Cityscapes (half resolution), MFNet, and ZJU datasets, while also having low complexity and being state-of-the-art in terms of speed among multimodal semantic segmentation models.

## 2. Related works

In this section, we provide a brief overview of previous single- and multi-modal semantic segmentation methods, given the high overlap between their proposed methods and techniques.

*2.1. Single-modal semantic segmentation*

The advent of the Fully Convolutional Network (FCN) [21] and the replacement of fully connected layers with convolutional layers marked a significant advancement in pixel-level classification. Similar to the FCN, the U-Net [22] and SegNet [23] adopted the encoder-decoder network architecture for semantic segmentation tasks. U-Net used skip connections to transfer the entire feature map, while SegNet transferred the max-pooling indices to reduce computational resources. To have a larger receptive field, PSPNet [24] proposed a Pyramid Pooling Module (PPM) that aggregates both local and global context information, while Deeplab [25], [26] used the Atrous Spatial Pyramid Pooling (ASPP) module to capture multi-scale context. Motivated by the integration of attention mechanisms into convolutional networks to capture feature dependencies [27], [28], semantic segmentation models such as DANet [29] and SANet [30] have effectively employed attention mechanisms. Building on these successes, transformer-based architectures have made remarkable progress in semantic segmentation. SETR [31] proposed an encoder with a transformer structure for the sequence-to-sequence prediction task. SegFormer [7] used a hierarchical encoder to achieve multi-scale representations. MaskFormer [32] approached semantic segmentation by predicting sets of masks rather than classifying each pixel individually.

In addition, some networks have improved towards real-time semantic segmentation. Specifically, ERFNet [33] used the residual connection and factorized convolution to reduce the computational cost while maintaining the accuracy; BiSeNetV1 [34] and BiSeNetV2 [35] used a dual-branch architecture for low- and high-level information; STDCSeg [20] proposed a Short-Term Dense Concatenate (STDC) backbone and a detail guidance module to guide the low-level layers to learn the spatial information; and PP-LiteSeg [36] proposed a lightweight decoder combined with the Unified Attention Fusion Module (UAFM) to strengthen the feature representations.

*2.2. Multi-modal semantic segmentation*

Despite significant advances in single-modal semantic segmentation, accuracy degrades under challenging conditions, especially when there is insufficient information in RGB images. Therefore, multimodal semantic segmentation networks have been widely studied to exploit complementary information with RGB images. The core of multimodal networks is the fusion strategy of RGB and complementary data. To perform cross-modal fusion, early works [11], [15], [37], [38] employed simple element-wise addition. More effectively, [3], [12], [18] used channel attention, and [16], [17], [19] used both channel and spatial attention modules for better fusion. To address the input noise, SA-Gate [4], CMX [5], and SpiderMesh [39] proposed the Separation-and-Aggregation Gate (SA-Gate), Cross-Modal Feature Rectification Module (CM-FRM), and Demand-guided Target Masking Module (DTM), respectively. In a simpler approach, LDFNet [40] concatenated the luminance and depth information for noise suppression. In addition, some multimodal semantic segmentation models have proposed specific approaches to improve their performance. For instance, ADSD [13] and CAINet [41] used auxiliary supervision with multiple decoders; IGFNet [42] used illumination to guide the fusion of RGB-T features; and CRM-RGBTSeg [43] improved the robustness and accuracy of RGB-T semantic segmentation by using a random masking strategy and self-distillation loss.

The majority of multimodal semantic segmentation models have focused on a particular type of input data. Only a few models, such as NLFNet [44] and CMX [5], have been proposed as RGB-X semantic segmentation networks. Good performance on one type of input data does not necessarily imply good performance in other domains. For example, RGB-D models such as ACNet [12] and SA-Gate [4] have low accuracy in the RGB-T semantic segmentation task [45]. Therefore, in this paper, we propose the CS-AFM module to effectively rectify and fuse different modalities. This module exhibits high generalization ability by considering cross-modal similarity. By using the CS-AFM module in the efficient encoder-decoder architecture, we propose CSFNet as a real-time RGB-X semantic segmentation model for driving applications.

## 3. The proposed method

In this section, we first describe the architecture of the proposed CSFNet, then introduce the Cosine Similarity Attention Fusion Module (CS-AFM) and the Efficient Context Module in detail.

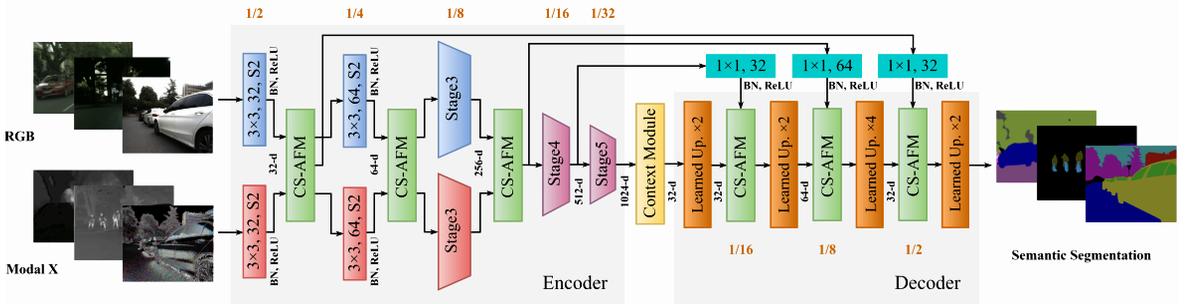

**Fig. 1.** Overview of our proposed CSFNet for real-time RGB-X semantic segmentation. The inputs are RGB and X (depth, thermal, or polarization) modalities. ($k_w \times k_h$, C, S2: Convolution with kernel size $k_w \times k_h$, C output channels and stride 2, BN: Batch Normalization and N-d: The channel dimension is N.)

### 3.1. Architecture overview

The overview of the proposed Cosine Similarity Fusion Network (CSFNet) is shown in Fig. 1. Our network follows an encoder-decoder structure optimized for real-time scene parsing. The encoder uses a double-branch network for the first three stages and a single-branch network for the last two stages. Using this structure with an efficient STDC [20] backbone reduces computational complexity and increases inference speed. Specifically, the RGB and X (depth, thermal, or polarization) modalities are effectively extracted and fused in the first three stages. In the last two stages, high-level features are extracted from the fused features. Due to the presence of noise in the input data and the importance of considering complementary features, we design the Cosine Similarity Attention Fusion Module (CS-AFM). This module uses cosine similarity in an attention-based approach to rectify and fuse the cross-modal features. Thus, we apply the CS-AFM module at the first three levels of the encoder network. To capture contextual information, we employ an efficient context module between the encoder and decoder. This module models the long-range dependencies using 1D convolutional layers. To improve efficiency, we leverage a lightweight decoder in CSFNet. Our decoder uses a combination of learned upsampling modules and CS-AFM modules for efficient feature recovery. The learned upsampling modules include a bilinear interpolation and 3×3 convolutional layers, followed by a batch normalization layer and the ReLU activation function. Except for the third learned upsampling module, which upsamples by a factor of 4 and uses two convolutional layers, the other three modules upsample by a factor of 2 and have only one convolutional layer. In addition, the three CS-AFM modules effectively fuse the skip connections and decoder feature maps. The skip connections reduce encoder feature maps with a 1×1 convolution to the same number of channels in the decoder.

*3.2. Cosine similarity attention fusion module*

Using global information for cross-modal feature fusion has been a common approach in previous research. Despite the improvements, global information cannot adequately discriminate between cross-modal features. To address this problem, we propose a novel Cosine Similarity Attention Fusion Module (CS-AFM) to rectify and fuse the input modalities in a more efficient way.

The structure of the CS-AFM module is shown in Fig. 2. Suppose $F_x \in \mathbb{R}^{C \times W \times H}$ and $F_y \in \mathbb{R}^{C \times W \times H}$ are the extracted feature maps of two encoder branches. In the CS-AFM module, the input features ($F_x$ and $F_y$) are downsampled using the adaptive average pooling layers. Then, cosine similarity is applied to measure the similarity between the two modalities. To perform the cosine similarity, the downsampled features are reshaped to dimension $C \times N$, where $N$ is equal to $P_w \times P_h$. The process of obtaining the similarity vector ($S_v$) from $\hat{F}_x$ and $\hat{F}_y$ is formulated as follows:

$$S_v = \frac{\hat{F}_x \cdot \hat{F}_y}{\|\hat{F}_x\| \times \|\hat{F}_y\|}, \tag{1}$$

where $S_v \in \mathbb{R}^{C \times 1}$ indicates the degree of similarity between the channels of the two input modalities, and its values range from -1 to 1.

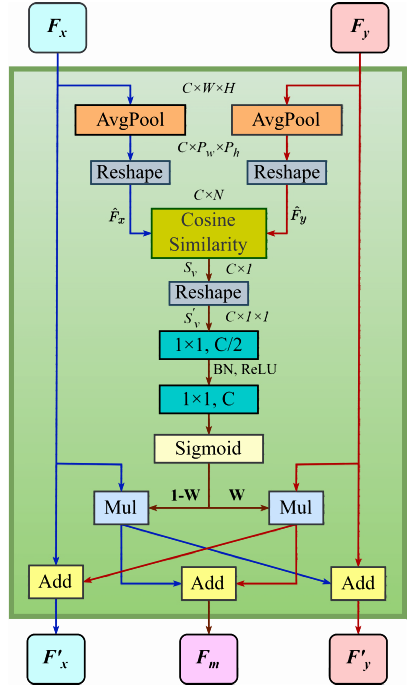

**Fig. 2.** The structure of our proposed Cosine Similarity Attention Fusion Module (CS-AFM).

To find the correlations between the channels and make an appropriate weight distribution, we employ two convolutional layers followed by a sigmoid function, as shown in the following equation:

$$W = \sigma\left(Conv_{1\times1}\left(ReLU\left(BN\left(Conv_{1\times1}(S'_v)\right)\right)\right)\right), \quad (2)$$

where $\sigma$, BN, and $S'_v$ denote the sigmoid function, batch normalization layer, and reshaped form of $S_v$, respectively.

To rectify the cross-modal features, the weights $W \in \mathbb{R}^{C\times1\times1}$ and $(1-W) \in \mathbb{R}^{C\times1\times1}$ are multiplied by their corresponding modalities and added to the other modality using the following equations:

$$\begin{aligned}F'_x &= F_x + F_y \times W, \\ F'_y &= F_y + F_x \times (1-W),\end{aligned} \quad (3)$$

where $F'_x \in \mathbb{R}^{C\times W\times H}$ and $F'_y \in \mathbb{R}^{C\times W\times H}$ are the rectified feature maps. In addition, to fuse the cross-modal features in the encoder and to fuse the skip connections and decoder features, weighted element-wise summation is used as the following equation:

$$F_m = F_y \times W + F_x \times (1-W), \quad (4)$$

where $F_m \in \mathbb{R}^{C\times W\times H}$ is the merged feature map.

*3.3. Efficient context module*

In order to achieve a richer semantic understanding, we propose an efficient context module to capture contextual information. As shown in Fig. 3, this module first applies an adaptive average pooling layer along with a 1×1 convolutional layer to reduce the dimension of the input feature maps from $C_{in} \times W/32 \times H/32$ to $C_{in}/4 \times S_w \times S_h$. Then, two 1D convolutional layers are used in two parallel branches to extract long-range dependencies. These layers use 4×1 and 1×4 kernels and reduce the channel dimension from $C_{in}/4$ to $C_{in}/16$. To merge the branches by element-wise summation, we leverage the bilinear interpolation method to upsample the features to the input resolution. Finally, a 3×3 convolutional layer is applied to reduce the channel dimension from $C_{in}/16$ to $C_{in}/32$. It is worth noting that all convolutional layers, except the last one, use a batch normalization layer and a ReLU activation function.

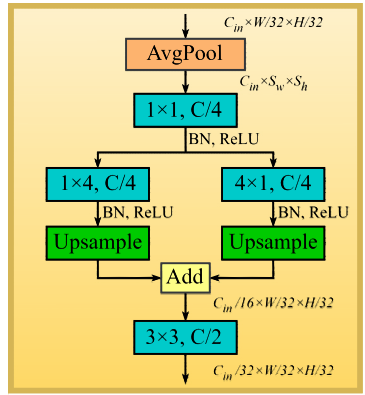

**Fig. 3.** Detailed structure of efficient context module.

# 4. Experiments and results

In this section, we first describe three benchmark datasets for RGB-D/T/P semantic segmentation tasks in driving scenarios. Then, we explain the implementation details of our proposed CSFNet and compare it with SOTA methods. Finally, we perform ablation experiments to demonstrate the effectiveness of our proposed method.

*4.1. Multimodal datasets*

To evaluate the performance of our proposed method, we use the Cityscapes [1], MFNet [2], and ZJU [3] datasets for the RGB-D, RGB-T and RGB-P semantic segmentation tasks, respectively.

**Cityscapes.** Cityscapes is an RGB-D dataset of urban street scenes. It consists of 5,000 finely annotated images with 19 semantic classes. The images have a resolution of 2048×1024 and are divided into 2,975 images for training, 500 for validation, and 1,525 for testing.

**MFNet.** The MFNet dataset contains 1569 RGB-T image pairs with a resolution of 640×480. It has 784 images for training, 392 for validation, and 393 for testing, provided in 9 semantic classes. There are 50% of daytime and 50% of nighttime images in the training set, while 25% of daytime and 25% of nighttime images are in the validation and test sets, respectively.

**ZJU.** The ZJU is an RGB-P dataset captured from complex campus street scenes and annotated in 8 semantic classes. It has 344 image pairs for training and 50 for validation. The original images have a resolution of 1224×1024, but are resized to 612×512. Each image pair contains four polarized images $[I_{0°}, I_{45°}, I_{90°}, I_{135°}]$, where $I_\alpha$ denotes the polarized image with polarization angle $\alpha$. In order to obtain the Angle of Linear Polarization (AoLP) representation, the Stokes vectors $S_1$ and $S_2$ can be derived from the following equation:

$$S_1 = I_{0°} - I_{90°},$$
$$S_2 = I_{45°} - I_{135°}, \qquad (5)$$

where $S_1$ and $S_2$ represent the ratio of $0°$ and $45°$ linear polarization over their perpendicular polarized portion. Then, AoLP can be formulated as:

$$AoLP = \frac{1}{2} arctan(S_1/S_2). \qquad (6)$$

*4.2. Implementation details*

We implement the proposed CSFNet using PyTorch 2.1.2, CUDA 12.1, CUDNN 8.9.0, and Python 3.10.13 on an NVIDIA RTX 3090 GPU (24 GB RAM). The STDC1 and STDC2 [20] backbones are pretrained on the ImageNet [46] dataset and employed on CSFNet-1 and CSFNet-2, respectively. For the Cityscapes dataset, the batch size, learning rate, and number of training epochs are set to 16, 0.02, and 300, while for the MFNet and ZJU datasets, they are set to 8, 0.01, and 600, respectively. The stochastic gradient descent (SGD) with momentum 0.9 and weight decay $5 \times 10^{-4}$ is adopted as the optimizer, and the polynomial learning rate policy with power 0.9 is used to reduce the learning rate during training.

For all datasets, random horizontal flipping, random scaling, random cropping, random color jittering, and normalization are used as data augmentation. The random scales are in the range [0.5, 1.75]. Except for Cityscapes, which is cropped to 1024×512, MFNet and ZJU use their image resolution as the crop size. We also concatenate the luminance and input depth information for noise suppression, similar to [40]. As mentioned, adaptive average pooling layers are applied in the proposed CS-AFM and efficient context modules. For more details, the output size of the adaptive average pooling layers for each dataset is shown in Table 1.

**Table 1**
The output size of the adaptive average pooling layers for each dataset in different modules (Lx means level x).

| Module | Cityscapes | MFNet & ZJU |
|---|---|---|
| CS-AFM-L1 | 32×16 | 24×16 |
| CS-AFM-L2 | 16×8 | 12×8 |
| CS-AFM-L3 | 8×4 | 6×4 |
| CS-AFM-L4 | 4×2 | 3×2 |
| Context Module | 8×4 | 5×5 |

*4.3. Comparison with SOTA models*

For a comprehensive comparison, we evaluate the proposed method on three RGB-D/T/P datasets using different metrics. The metrics include the number of parameters, frames per second (FPS), floating point operations per second (FLOPs), and mean intersection over union (mIoU).

**Table 2**
Comparison with state-of-the-art methods on Cityscapes *val* set at half resolution (1024×512).

| Method | Modal | Backbone | GPU | Params | FPS | mIoU |
|---|---|---|---|---|---|---|
| SwiftNet [47] | RGB | ResNet18 | GTX 1080Ti | 11.8 | 134.9 | 70.2 |
| BiSeNetV2 [35] | RGB | - | GTX 1080Ti | - | 156 | 73.4 |
| BiSeNetV2-L [35] | RGB | - | GTX 1080Ti | - | 47.3 | 75.8 |
| STDC1-Seg50* [20] | RGB | STDC1 | RTX 3090 | - | 145.6 | 72.2 |
| STDC2-Seg50* [20] | RGB | STDC2 | RTX 3090 | - | 96.16 | 74.2 |
| PP-LiteSeg-T1* [36] | RGB | STDC1 | RTX 3090 | - | **166.4** | 73.1 |
| PP-LiteSeg-B1* [36] | RGB | STDC2 | RTX 3090 | - | 105.8 | 75.3 |
| LDFNet* [40] | RGB-D | ERFNet | RTX 3090 | **2.31** | 68.5 | 68.48 |
| ESANet* [18] | RGB-D | R18-NBt1D | RTX 3090 | - | 70.22 | 74.65 |
| ESANet* [18] | RGB-D | R34-NBt1D | RTX 3090 | - | 43.63 | 75.22 |
| SGACNet* [17] | RGB-D | R18-NBt1D | RTX 3090 | 22.1 | 50.1 | 73.3 |
| SGACNet* [17] | RGB-D | R34-NBt1D | RTX 3090 | 35.6 | 35.7 | 74.1 |
| CSFNet-1 | RGB-D | STDC1 | RTX 3090 | 11.31 | 106.1 | 74.73 |
| CSFNet-2 | RGB-D | STDC2 | RTX 3090 | 19.37 | 72.3 | **76.36** |

\* The inference speed of marked models is retested on an NVIDIA RTX 3090 GPU.

*1) Comparison results on Cityscapes:* Given the importance of the speed metric, Table 2 presents the results on the Cityscapes dataset at half resolution (1024×512). According to the results, CSFNet-2 achieves mIoU of 76.36%, which is the highest accuracy among both RGB and RGB-D semantic segmentation models. It also attains a speed of 72.3 FPS, which, after its lighter version (CSFNet-1) with 106.1 FPS, has the highest speed among the RGB-D models. Despite the improvement in speed, the proposed CSFNet doesn't reach the speed of the fastest RGB models due to the higher complexity of RGB-D processing. Remarkably, both CSFNet-1 and CSFNet-2 have fewer parameters and higher accuracy than the efficient SGACNet model [17], which shows the high efficiency of the proposed architecture. Fig. 4 presents some of the qualitative results obtained on the CityScapes dataset, showcasing the remarkable semantic segmentation performance of our proposed method.

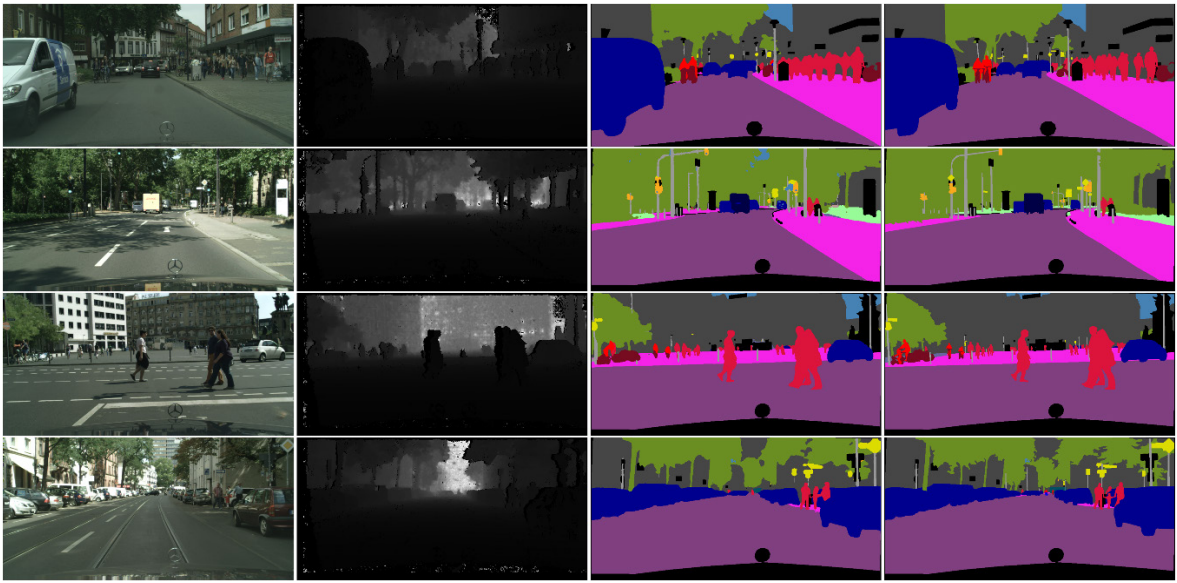

**Fig. 4.** Visual results of CSFNet on the Cityscapes *val* set (half resolution). From left to right: RGB input, depth input, prediction, and ground truth.

*2) Comparison results on MFNet:* Among the methods compared in Table 3, CSFNet-2 achieves the second-highest mIoU of 59.98%, surpassed only by the SOTA method CRM-RGBTSeg [43] with mIoU of 61.4%. It also achieves the highest IoU of 26.05% for the Guardrail class. Our proposed method has remarkable results in terms of computational efficiency. Specifically, CSFNet-1 achieves mIoU of 56.05% with only 11.30M parameters and 27.17G FLOPs. As presented in Table 4, CSFNet-1 with 106.3 FPS and CSFNet-2 with 72.7 FPS are SOTA in terms of inference speed, and they also have the highest accuracy among all real-time RGB-T semantic segmentation methods.

According to the results in Table 5, CSFNet-1 achieves the highest accuracy with mIoU of 55.27% for the daytime scenes. In spite of weak illumination and noisy information in nighttime RGB images, CSFNet-2 demonstrates superior performance, attaining the highest mIoU of 60.80%. These results show the effective performance of the CS-AFM module in fusing and rectifying cross-modal features. The visual results for daytime and nighttime scenarios of the MFNet dataset are shown in Fig. 5. It can be observed that the CSFNet has accurate and detailed segmentation predictions, particularly in poor lighting conditions.

**Table 3**
Comparison with state-of-the-art methods on the MFNet dataset. The best results are shown in bold font.

| Method | Backbone | Params | FLOPs | Car | Person | Bike | Curve | Car Stop | Guardrail | Color Cone | Bump | mIoU |
|---|---|---|---|---|---|---|---|---|---|---|---|---|
| MFNet [2] | - | **0.73** | - | 65.9 | 58.9 | 42.9 | 29.9 | 9.9 | 8.5 | 25.2 | 27.7 | 39.7 |
| RTFNet [11] | ResNet50 | 185.24 | 245.71 | 86.3 | 67.8 | 58.2 | 43.7 | 24.3 | 3.6 | 26.0 | 57.2 | 51.7 |
| RTFNet [11] | ResNet152 | 254.51 | 337.04 | 87.4 | 70.3 | 62.7 | 45.3 | 29.8 | 0.0 | 29.1 | 55.7 | 53.2 |
| FuseSeg [38] | DenseNet161 | 141.52 | 193.40 | 87.9 | 71.7 | 64.6 | 44.8 | 22.7 | 6.4 | 46.9 | 47.9 | 54.5 |
| NLFNet [44] | ResNet-18 | - | - | 88.5 | 69.0 | 63.9 | 47.8 | 25.6 | 6.1 | 45.0 | 44.7 | 54.3 |
| FEANet [16] | ResNet152 | - | - | 87.8 | 71.1 | 61.1 | 46.5 | 22.1 | 6.6 | **55.3** | 48.9 | 55.3 |
| ABMDRNet [45] | ResNet50 | 64.60 | 194.33 | 84.8 | 69.6 | 60.3 | 45.1 | 33.1 | 5.1 | 47.4 | 50 | 54.8 |
| EAEFNet [19] | ResNet152 | 200.4 | 147.3 | 87.6 | 72.6 | 68.3 | 48.6 | 35.0 | 14.2 | 52.4 | 58.3 | 58.9 |
| CAINet [41] | MobileNet-V2 | 12.16 | 123.62 | 88.5 | 66.3 | **68.7** | 55.4 | 31.5 | 9.0 | 48.9 | **60.7** | 58.6 |
| SpiderMesh [39] | MiT-B4 | - | 398.8 | 89.9 | **75.3** | 64.8 | 51.5 | 31.4 | 4.5 | 54.5 | 55.9 | 58.4 |
| IGFNet [42] | MiT-B2 | 67.44 | 69.18 | 88.0 | 74.0 | 62.7 | 48.2 | 36.0 | 14.2 | 52.4 | 57.5 | 59.0 |
| CMX [5] | MiT-B2 | - | - | 89.4 | 74.8 | 64.7 | 47.3 | 30.1 | 8.1 | 52.4 | 59.4 | 58.2 |
| CMX [5] | MiT-B4 | - | - | **90.1** | 75.2 | 64.5 | 50.2 | 35.3 | 8.5 | 54.2 | 60.6 | 59.7 |
| CRM-RGBTSeg [43] | Swin-B | - | - | 90.0 | 75.1 | 67.0 | 45.2 | **49.7** | 18.4 | 54.2 | 54.4 | **61.4** |
| CSFNet-1 | STDC1 | 11.30 | **27.17** | 86.91 | 71.35 | 62.21 | 44.16 | 33.10 | 7.96 | 46.11 | 54.54 | 56.05 |
| CSFNet-2 | STDC2 | 19.36 | 47.82 | 87.36 | 73.59 | 63.74 | 49.28 | 43.49 | 26.05 | 47.32 | 50.79 | 59.98 |

**Table 4**
The inference speed of networks on the MFNet dataset.

| Method | GPU | FPS | mIoU |
|---|---|---|---|
| MFNet [2] | GTX Titan X | 55.6 | 39.7 |
| RTFNet-50* [11] | RTX 3090 | 40.88 | 51.7 |
| RTFNet-152* [11] | RTX 3090 | 25.24 | 53.2 |
| FuseSeg-161 [38] | RTX 2080 Ti | 30.01 | 54.5 |
| NLFNet [44] | GTX 1080Ti | 35.6 | 54.3 |
| FEANet* [16] | RTX 3090 | 21.13 | 55.3 |
| CSFNet-1 | RTX 3090 | **106.3** | 56.05 |
| CSFNet-2 | RTX 3090 | 72.7 | **59.98** |

* The inference speed of marked models is retested on an NVIDIA RTX 3090 GPU.

**Table 5**
The comparative results for the daytime and nighttime scenarios on the MFNet dataset.

| Method | Backbone | mIoU (%) Daytime | mIoU (%) Nighttime |
|---|---|---|---|
| RTFNet [11] | ResNet152 | 45.8 | 54.8 |
| FuseSeg [38] | DenseNet161 | 47.8 | 54.6 |
| NLFNet [44] | ResNet-18 | 50.3 | 54.8 |
| SpiderMesh [39] | ResNet152 | 52.0 | 56.0 |
| CMX [5] | MiT-B2 | 51.3 | 57.8 |
| CMX [5] | MiT-B4 | 52.5 | 59.4 |
| CSFNet-1 | STDC1 | **55.27** | 46.45 |
| CSFNet-2 | STDC2 | 49.34 | **60.80** |

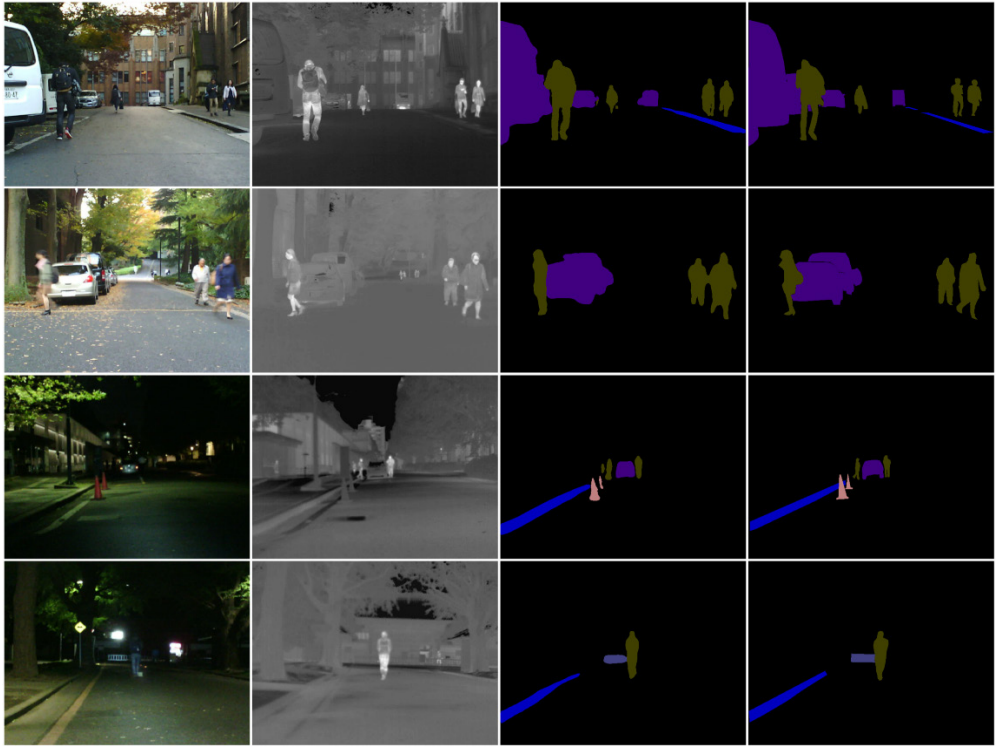

**Fig. 5.** Visual results of CSFNet on the MFNet dataset. From left to right: RGB input, thermal input, prediction, and ground truth.

*3) Comparison results on ZJU:* In Table 6, we compare our proposed CSFNet with SOTA RGB and RGB-P semantic segmentation methods using the ZJU dataset. According to the results, CSFNet-2 and CSFNet-1 achieve mIoUs of 91.40% and 90.85%, respectively, which are the third- and fourth-highest accuracies after CMX-B4 [5] and CMX-B2 [5]. Despite the competitive results in terms of accuracy, CSFNet-1 with 108.5 FPS and CSFNet-2 with 75 FPS have the highest inference speed among all RGB-P semantic segmentation models. The qualitative results of the CSFNet on the ZJU dataset are illustrated in Fig. 6.

Table 6
Comparison with state-of-the-art methods on the ZJU dataset.

| Method | Modal | Backbone | Building | Glass | Car | Road | Tree | Sky | Pedestrian | Bicycle | mIoU |
|---|---|---|---|---|---|---|---|---|---|---|---|
| SegFormer [7] | RGB | MiT-B2 | 90.6 | 79.0 | 92.8 | 96.6 | 96.2 | 89.6 | 82.9 | 89.3 | 89.6 |
| NLFNet [44] | RGB-AoLP | ResNet-18 | 85.4 | 77.1 | 93.5 | 97.7 | 93.2 | 85.9 | 56.9 | 85.5 | 84.4 |
| EAFNet [3] | RGB-AoLP | ResNet-18 | 87.0 | 79.3 | 93.6 | 97.4 | 95.3 | 87.1 | 60.4 | 85.6 | 85.7 |
| CMX [5] | RGB-AoLP | MiT-B2 | 91.5 | 87.3 | 95.8 | 98.2 | 96.6 | 89.3 | 85.6 | 91.9 | 92.0 |
| CMX [5] | RGB-AoLP | MiT-B4 | 91.6 | 88.8 | 96.3 | 98.3 | 96.8 | 89.7 | 86.2 | 92.8 | 92.6 |
| CSFNet-1 | RGB-AoLP | STDC1 | 90.19 | 84.75 | 95.21 | 97.91 | 96.23 | 89.18 | 82.49 | 90.86 | 90.85 |
| CSFNet-2 | RGB-AoLP | STDC2 | 91.10 | 86.11 | 95.54 | 98.03 | 96.51 | 89.50 | 83.18 | 91.26 | 91.40 |

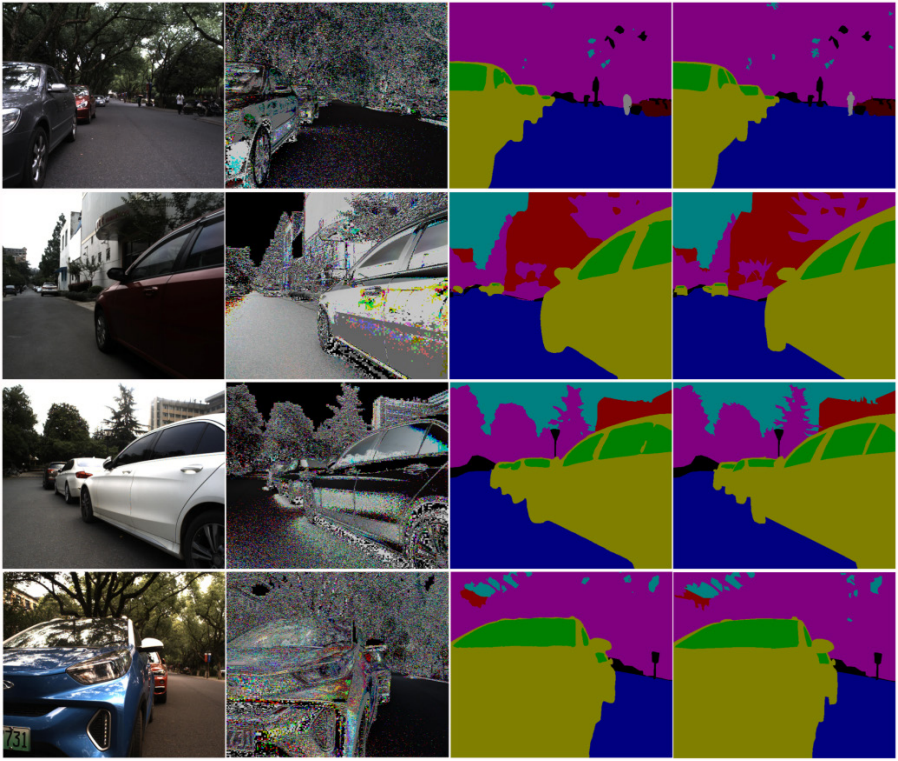

**Fig. 6.** Visual results of CSFNet on the ZJU dataset. From left to right: RGB input, AoLP input, prediction, and ground truth.

## 4.4. Ablation Study

We performed ablation experiments on the Cityscapes *val* dataset to verify the effectiveness of the proposed CSFNet. As shown in Table 7, we evaluate the performance of CSFNet-1 with different levels of parallelization in the encoder network. According to the results, adding two parallel stages (from 3 to 5) leads to an 11.58% increase in the number of parameters, an 83.77% increase in FLOPs, and a 29.87% decrease in inference speed. Despite this performance degradation, mIoU accuracy improved by only 0.07%. This slight improvement in accuracy demonstrates the strength of the CS-AFM module in fusing and rectifying cross-modal features in the first three stages. In other words, by improving the fusion of cross-modal features at lower levels, CS-AFM paves the way for the use of a single-branch network at higher levels.

**Table 7**
Performance comparison of CSFNet-1 at different levels of parallelization using the Cityscapes *val* set (half resolution).

| Method   | Dual-Branch | Params | FLOPs | FPS   | mIoU  |
|----------|-------------|--------|-------|-------|-------|
| CSFNet-1 | 3 stages    | **11.31** | **47.28** | **106.1** | 74.73 |
| CSFNet-1 | 4 stages    | 11.57  | 67.11 | 82.9  | 74.75 |
| CSFNet-1 | 5 stages    | 12.62  | 86.89 | 74.4  | **74.80** |

We also evaluate the effectiveness of using the CS-AFM module in the decoder network. As shown in Table 8, using the CS-AFM module in the decoder achieves higher accuracy with more parameters and a lower FPS compared to applying element-wise addition.

**Table 8**
Performance evaluation of the CS-AFM module in the decoder network using the Cityscapes *val* set (half resolution).

| Method | CS-AFM in decoder | Params | FPS | mIoU |
| --- | --- | --- | --- | --- |
| CSFNet-1 | ✓ | 11.31 | 106.1 | **74.73** |
| CSFNet-1 |  | **11.30** | **111.2** | 74.28 |
| CSFNet-2 | ✓ | 19.37 | 72.3 | **76.36** |
| CSFNet-2 |  | **19.36** | **77.1** | 76.01 |

## 5. Conclusion

Considering the critical need for real-time processing in driving applications and to overcome the speed limitations of existing multimodal semantic segmentation methods, we design CSFNet, a real-time RGB-X semantic segmentation model. The key component of this model is the proposed CS-AFM module. The CS-AFM module rectifies and fuses the input modalities by employing the cosine similarity between the corresponding channels in an attention-based approach. Given the high generalization ability of the CS-AFM module and its effective performance in low-level feature fusion, we propose an optimized encoder, followed by an effective context module and a lightweight decoder. Experiments on three RGB-D, RGB-T, and RGB-P datasets demonstrate that our CSFNet model achieves high efficiency and is the fastest multimodal semantic segmentation model while maintaining competitive accuracy with SOTA methods. The combination of these advantages makes the CSFNet model ideal for real-time applications in autonomous driving and robotics and facilitates its implementation on embedded hardware.

## References

[1] M. Cordts *et al.*, "The Cityscapes Dataset for Semantic Urban Scene Understanding," *in Proc. 2016 IEEE Conf. Comput. Vis. Pattern Recognit. (CVPR)*, Jun. 2016, pp. 3213-3223.
[2] Q. Ha, K. Watanabe, T. Karasawa, Y. Ushiku, and T. Harada, "MFNet:Towards real-time semantic segmentation for autonomous vehicles with multi-spectral scenes," *in Proc. IROS*, Sep. 2017, pp. 5108–5115.
[3] K. Xiang, K. Yang, and K. Wang, "Polarization-driven semantic segmentation via efficient attention-bridged fusion," *Opt. Exp.*, vol. 29, no. 4, pp. 4802–4820, 2021.
[4] X. Chen et al., "Bi-directional cross-modality feature propagation with separation-and-aggregation gate for RGB-D semantic segmentation," *in Proc. ECCV*, 2020, pp. 561–577.
[5] J. Zhang, H. Liu, K. Yang, X. Hu, R. Liu and R. Stiefelhagen, "CMX: Cross-Modal Fusion for RGB-X Semantic Segmentation With Transformers," *IEEE Trans. Intell. Transp. Syst*, vol. 24, no. 12, pp. 14679-14694, Dec. 2023.
[6] L.-C. Chen, Y. Zhu, G. Papandreou, F. Schroff, and H. Adam, "Encoder–decoder with atrous separable convolution for semantic image segmentation," *in Proc. ECCV*, 2018, pp. 801–818.
[7] E. Xie, W. Wang, Z. Yu, A. Anandkumar, J. M. Alvarez, and P. Luo, "SegFormer: Simple and efficient design for semantic segmentation with transformers," *in Proc. NeurIPS*, 2021, pp. 12077–12090.
[8] Couprie, Camille, Clément Farabet, Laurent Najman, and Yann LeCun, "Indoor semantic segmentation using depth information." 2013, *arXiv:1301.3572*.
[9] Wang, Jinghua, Zhenhua Wang, Dacheng Tao, Simon See, and Gang Wang, "Learning common and specific features for RGB-D semantic segmentation with deconvolutional networks," *in Proc. Comp. Vision–ECCV 2016: 14th Euro. Conf.*, 2016, pp. 664-679.

[10] Y. Cheng, R. Cai, Z. Li, X. Zhao, and K. Huang, "Locality-sensitive deconvolution networks with gated fusion for RGB-D indoor semantic segmentation," in *Proc. IEEE Conf. Comput. Vis. Pattern Recognit. (CVPR)*, Jul. 2017, pp. 3029–3037.

[11] Y. Sun, W. Zuo, and M. Liu, "RTFNet: RGB-thermal fusion network for semantic segmentation of urban scenes," *IEEE Robot. Autom. Lett.*, vol. 4, no. 3, pp. 2576–2583, Jul. 2019.

[12] X. Hu, K. Yang, L. Fei, and K. Wang, "ACNET: Attention based network to exploit complementary features for RGBD semantic segmentation," in *Proc. ICIP*, Sep. 2019, pp. 1440–1444.

[13] Zhang, Yang, Yang Yang, Chenyun Xiong, Guodong Sun, and Yanwen Guo, "Attention-based dual supervised decoder for RGBD semantic segmentation." 2022, *arXiv:2201.01427*.

[14] Zhang, H., Sheng, V.S., Xi, X., Cui, Z. and Rong, H., "Overview of RGBD semantic segmentation based on deep learning," *Journ. Ambient Intell. Hum. Comp.*, pp.13627-13645, Apr. 2023.

[15] C. Hazirbas, L. Ma, C. Domokos, and D. Cremers, "FuseNet: Incorporating depth into semantic segmentation via fusion-based CNN architecture," in *Proc. ACCV*, 2016, pp. 213–228.

[16] F. Deng et al., "FEANet: Feature-enhanced attention network for RGBthermal real-time semantic segmentation," in *Proc. IROS*, Sep. 2021, pp. 4467–4473.

[17] Y. Zhang, C. Xiong, J. Liu, X. Ye and G. Sun, "Spatial Information-Guided Adaptive Context-Aware Network for Efficient RGB-D Semantic Segmentation," *IEEE Sensors Journal*, vol. 23, no. 19, pp. 23512-23521, 1 Oct. 2023,

[18] D. Seichter, M. Kohler, B. Lewandowski, T. Wengefeld, and H.-M. Gross, "Efficient RGB-D semantic segmentation for indoor scene analysis," in *Proc. ICRA*, May 2021, pp. 13525–13531.

[19] M. Liang, J. Hu, C. Bao, H. Feng, F. Deng and T. L. Lam, "Explicit Attention-Enhanced Fusion for RGB-Thermal Perception Tasks," *IEEE Robot. Auto. Lett.*, vol. 8, no. 7, pp. 4060-4067, July 2023.

[20] Fan, Mingyuan, Shenqi Lai, Junshi Huang, Xiaoming Wei, Zhenhua Chai, Junfeng Luo, and Xiaolin Wei. "Rethinking bisenet for real-time semantic segmentation." in *Proc. of the IEEE/CVF conf. on comput. Vis. pattern recognit.*, 2021, pp. 9716-9725.

[21] J. Long, E. Shelhamer, and T. Darrell, "Fully convolutional networks for semantic segmentation," in *Proc. IEEE/CVF Conf. Comput. Vis. Pattern Recognit.*, 2015, pp. 3431–3440.

[22] O. Ronneberger, P. Fischer, and T. Brox, "U-Net convolutional networks for biomedical image segmentation," in *Proc. Int. Conf. Med. Image Comput. Comput.-Assist. Interv.*, 2015, pp. 234–241.

[23] V. Badrinarayanan, A. Kendall and R. Cipolla, "SegNet: A Deep Convolutional Encoder-Decoder Architecture for Image Segmentation," *IEEE Trans Patt Anal Mach Intell*, vol. 39, no. 12, pp. 2481-2495, 1 Dec. 2017.

[24] H. Zhao, J. Shi, X. Qi, X. Wang, and J. Jia, "Pyramid scene parsing network," in *Proc. IEEE Conf. Comput. Vis. Pattern Recognit. (CVPR)*, Jul. 2017, pp. 2881–2890.

[25] L.-C. Chen, G. Papandreou, I. Kokkinos, K. Murphy, and A. L. Yuille, "DeepLab: Semantic image segmentation with deep convolutional nets, atrous convolution, and fully connected CRFs," *IEEE Trans. Pattern Anal. Mach. Intell.*, vol. 40, no. 4, pp. 834–848, Apr. 2018.

[26] Chen, Liang-Chieh, George Papandreou, Florian Schroff, and Hartwig Adam, "Rethinking atrous convolution for semantic image segmentation." 2017, *arXiv:1706.05587*.

[27] X. Wang, R. Girshick, A. Gupta, and K. He, "Non-local neural networks," in *Proc. IEEE/CVF Conf. Comput. Vis. Pattern Recognit.*, Jun. 2018, pp. 7794–7803.

[28] J. Hu, L. Shen, and G. Sun, "Squeeze-and-excitation networks," *IEEE/CVF Conf. Compu. Vis. Pattern Recognit. (CVPR)*, 2018, pp. 7132–7141.

[29] Zhong, Zilong, Zhong Qiu Lin, Rene Bidart, Xiaodan Hu, Ibrahim Ben Daya, Zhifeng Li, Wei-Shi Zheng, Jonathan Li, and Alexander Wong. "Squeeze-and-attention networks for semantic segmentation." in *Proc. IEEE/CVF conf. comput. Vis. pattern recognit.*, 2020, pp. 13065-13074.

[30] J. Fu et al., "Dual attention network for scene segmentation," in *Proc. CVPR*, Jun. 2019, pp. 3146–3154.

[31] S. Zheng et al., "Rethinking semantic segmentation from a sequenceto- sequence perspective with transformers," in *Proc. CVPR*, Jun. 2021, pp. 6881–6890.

[32] B. Cheng, A. Schwing, and A. Kirillov, "Per-pixel classification is not all you need for semantic segmentation," *Proc. NeurIPS*, vol. 34, pp. 17 864–17875, Oct. 2021.

[33] Romera, Eduardo, José M. Alvarez, Luis M. Bergasa, and Roberto Arroyo. "Erfnet: Efficient residual factorized convnet for real-time semantic segmentation." *IEEE Trans. Intell. Transp. Sys.*, no. 1 pp.263-272, 2017.

[34] Yu, Changqian, Jingbo Wang, Chao Peng, Changxin Gao, Gang Yu, and Nong Sang. "Bisenet: Bilateral segmentation network for real-time semantic segmentation." *In Proc. Eur. Conf. Compu. Vis. (ECCV)*, 2018, pp. 325-341.

[35] Yu, Changqian, Changxin Gao, Jingbo Wang, Gang Yu, Chunhua Shen, and Nong Sang. "Bisenet v2: Bilateral network with guided aggregation for real-time semantic segmentation." *Int. Jour. Compu. Vis.*, pp.3051-3068, sep. 2021.

[36] Peng, Juncai, Yi Liu, Shiyu Tang, Yuying Hao, Lutao Chu, Guowei Chen, Zewu Wu et al. "Pp-liteseg: A superior real-time semantic segmentation model." 2022, *arXiv:2204.02681*.

[37] Jiang, Jindong, Lunan Zheng, Fei Luo, and Zhijun Zhang, "Rednet: Residual encoder-decoder network for indoor rgb-d semantic segmentation." 2018, *arXiv:1806.01054*.

[38] Y. Sun, W. Zuo, P. Yun, H. Wang, and M. Liu, "FuseSeg: Semantic segmentation of urban scenes based on RGB and thermal data fusion," *IEEE Trans. Autom. Sci. Eng.*, vol. 18, no. 3, pp. 1000–1011, Jul. 2021.

[39] Fan, Siqi, Zhe Wang, Yan Wang, and Jingjing Liu, "Spidermesh: Spatial-aware demand-guided recursive meshing for rgb-t semantic segmentation." 2023, *arXiv:2303.08692*.

[40] S.-W. Hung, S.-Y. Lo, and H.-M. Hang, "Incorporating luminance, depth and color information by a fusion-based network for semantic segmentation," *in Proc IEEE Int. Conf. Image Proc*, 2019, pp. 2374–2378.

[41] Y. Lv, Z. Liu and G. Li, "Context-Aware Interaction Network for RGB-T Semantic Segmentation," *IEEE Trans. Multi.*, vol. 26, pp. 6348-6360, Jan. 2024.

[42] H. Li and Y. Sun, "IGFNet: Illumination-Guided Fusion Network for Semantic Scene Understanding using RGB-Thermal Images," *in Proc IEEE Int. Conf. Robot. Biom. (ROBIO)*, Dec. 2023, pp. 1-6.

[43] Shin, Ukcheol, Kyunghyun Lee, In So Kweon, and Jean Oh. "Complementary random masking for RGB-thermal semantic segmentation." 2023, *arXiv:2303.17386*.

[44] Yan, Ran, Kailun Yang, and Kaiwei Wang. "NLFNet: Non-local fusion towards generalized multimodal semantic segmentation across RGB-depth, polarization, and thermal images." *in Proc IEEE int. conf. robot. Biom. (ROBIO)*, Dec. 2021, pp. 1129-1135.

[45] Zhang, Qiang, Shenlu Zhao, Yongjiang Luo, Dingwen Zhang, Nianchang Huang, and Jungong Han. "ABMDRNet: Adaptive-weighted bi-directional modality difference reduction network for RGB-T semantic segmentation." *in Proc. IEEE/CVF Conf. Compu. Vis. Pattern Recognit.*, 2021, pp. 2633-2642.

[46] Deng, Jia, Wei Dong, Richard Socher, Li-Jia Li, Kai Li, and Li Fei-Fei. "Imagenet: A large-scale hierarchical image database." *in Proc. IEEE conf. compu. Vis. pattern recognit.*, 2009, pp. 248-255.

[47] M. Oršic, I. Krešo, P. Bevandic, and S. Šegvic, "In defense of pretrained ImageNet architectures for real-time semantic segmentation of road-driving images," *in Proc. CVPR*, Jun. 2019, pp. 12607–12616.